\documentclass[letterpaper, 10 pt, conference]{ieeeconf}  % Comment this line out if you need a4paper

\IEEEoverridecommandlockouts                              % This command is only needed if 
\usepackage{graphicx} % for pdf, bitmapped graphics files
\usepackage{amsmath} % assumes amsmath package installed
\usepackage{amssymb}  % assumes amsmath package installed
\usepackage{subfig}
\usepackage{svg}
\usepackage{booktabs}
\usepackage[american]{babel}
\usepackage{array}
\usepackage{multirow}
\newcolumntype{P}[1]{>{\centering\arraybackslash}p{#1}}
\DeclareMathOperator*{\argmin}{arg\,min}
\usepackage{algorithm}
\usepackage[noend]{algpseudocode}
\let\oldReturn\Return
\renewcommand{\Return}{\State\oldReturn}
\usepackage{algorithm,algpseudocode}
\makeatletter
\newcommand{\StatexIndent}[1][3]{%
  \setlength\@tempdima{\algorithmicindent}%
  \Statex\hskip\dimexpr#1\@tempdima\relax}
\title{\LARGE \bf
Search-based Path Planning for a High Dimensional Manipulator in Cluttered Environments Using Optimization-based Primitives}

\author{Muhammad Suhail Saleem$^{1}$, Raghav Sood$^{1}$, Sho Onodera$^{2}$, Rohit Arora$^{2}$,\\ Hiroyuki Kanazawa$^{2}$, and Maxim Likhachev$^{1}$% <-this % stops a space
\thanks{$^{1}$ M.S. Saleem, R. Sood, and M. Likhachev are with the Robotics Institute, Carnegie Mellon University, Pittsburgh, PA 15213, USA. {\small e-mail: \tt \{msaleem2, raghavso, mlikhach\}@andrew.cmu.edu}.}%
\thanks{$^{2}$ S. Onodera, R. Arora, and H. Kanazawa are with the Machinery
Research Department, Research and Innovation Center, Mitsubishi Heavy
Industries, Takasago, Hyogo, Japan.}
}

\begin{document}

\maketitle
\thispagestyle{empty}
\pagestyle{empty}

%%%%%%%%%%%%%%%%%%%%%%%%%%%%%%%%%%%%%%%%%%%%%%%%%%%%%%%%%%%%%%%%%%%%%%%%%%%%%%%%
\begin{abstract}

In this work we tackle the path planning problem for a 21-dimensional snake robot-like manipulator, navigating a cluttered gas turbine for the purposes of inspection. Heuristic search based approaches are effective planning strategies for common manipulation domains.  However, their performance on high dimensional systems is heavily reliant on the effectiveness of the action space and the heuristics chosen. The complex nature of our system, reachability constraints, and highly cluttered turbine environment renders naive choices of action spaces and heuristics ineffective. To this extent we have developed i) a methodology for dynamically generating actions based on online optimization that help the robot navigate narrow spaces, ii) a technique for lazily generating these computationally expensive optimization actions to effectively utilize resources, and iii) heuristics that reason about the homotopy classes induced by the blades of the turbine in the robot workspace and a Multi-Heuristic framework which guides the search along the relevant classes. The impact of our contributions is presented through an experimental study in simulation, where the 21 DOF manipulator navigates towards regions of inspection within a turbine.

\end{abstract}

%%%%%%%%%%%%%%%%%%%%%%%%%%%%%%%%%%%%%%%%%%%%%%%%%%%%%%%%%%%%%%%%%%%%%%%%%%%%%%%%
\section{INTRODUCTION}
Highly articulated robotic systems have taken several forms over the years - interplanetary and lunar rovers, serpentine robots, and humanoid robots to name a few. Their development has been motivated by the superior mobility they offer when compared to lower dimensional systems. Their capacity for complex and expressive maneuvers have made them attractive choices in several domains including unknown/complex terrain navigation, search and rescue, and inspection \cite{snakeFighter}\cite{snakeInspection}. While their dimensionality is responsible for their superior mobility, it is the same which makes the tasks of planning for and controlling them difficult.

In this paper, we will focus on developing a path planning algorithm for a 21-DOF snake robot-like manipulator navigating a highly cluttered gas turbine towards regions of inspection. The curse of dimensionality severely cripples the performance of planning algorithms in our problem domain. This is further exacerbated by the extremely constrained environment created by the turbine. Further, even the simple yet essential operations of computing forward kinematics, inverse kinematics, and collision checking for our system are computationally expensive. These reasons call for an intelligent planning algorithm that utilizes domain knowledge to compute feasible paths in a reasonable amount of time. 

\begin{figure}[t]
    \centering
    \includegraphics[width = 0.7\columnwidth]{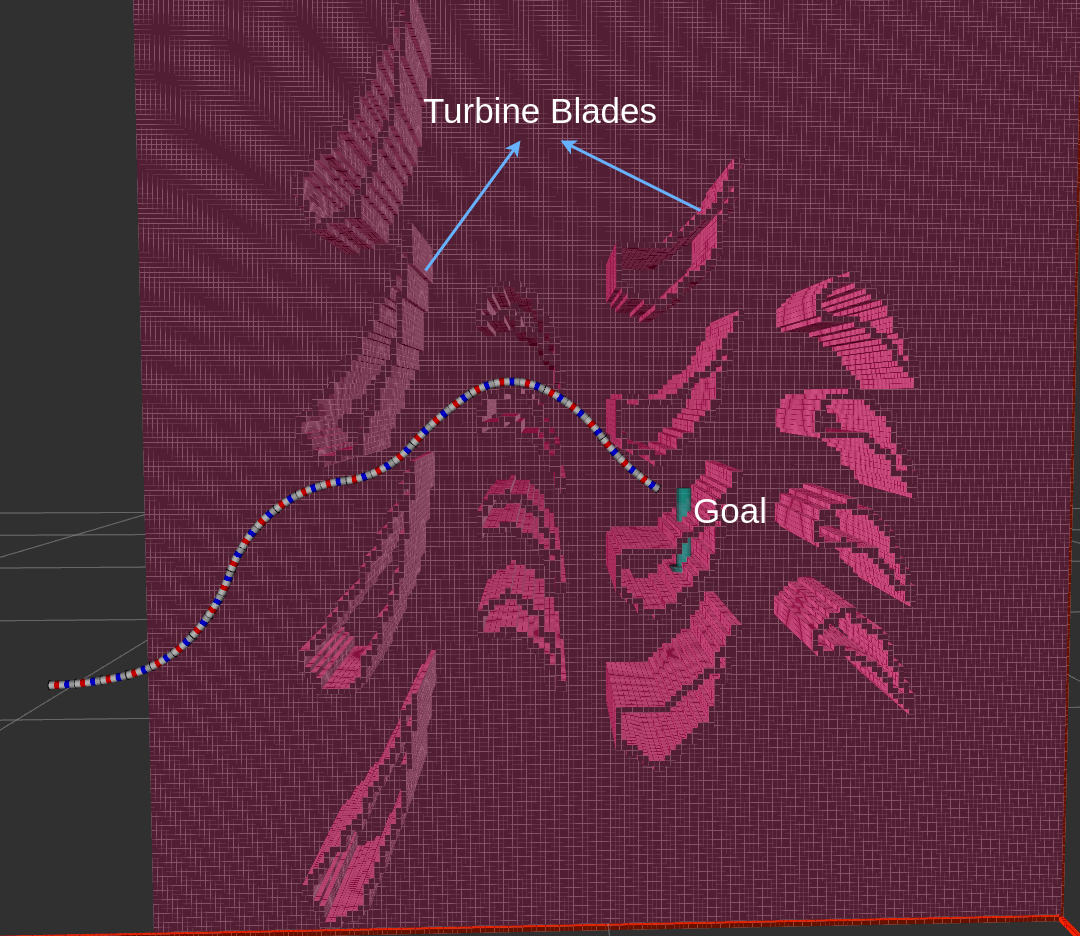}
    \caption{Snake robot-like manipulator navigating towards an inspection pose within the turbine (indicated by the cyan arrow).}
    \label{fig:intro} \vspace{-0.5cm}
\end{figure}

% Before we proceed, we would like to draw the distinction between the robot under discussion and a snake robot. 
The overall design of the robot we focus on, as can be seen in Fig. \ref{fig:intro}. (more details of the design and structure can be found in Section II A) is very similar to a snake robot's design. They share properties like modularity, small cross-section to length ratio, and hyper-redundancy. However, the major distinction between the two stems from the fact that the robot under discussion is not self-propelling. There is an external prismatic actuator that propels the robot. In the past, the term snake robot has been used in the context of self-propelling robots, i.e., robots which locomote purely by changing their shape. This aspect creates significant challenges from the standpoint of controls and there are several works dedicated towards this problem \cite{snakeRobotBook} \cite{snakeDesignAndControl}. Since our system is externally propelled while having a design very similar to a snake robot's, we will refer to our system as a snake robot-like manipulator.

The three major classes of planning strategies that are generally adopted for manipulation tasks include - i) Sampling-based strategies \cite{RRT}\cite{RRT-Connect}\cite{PRM} ii) Trajectory Optimization techniques \cite{CHOMP}\cite{STOMP}, and iii) Search-based planning approaches \cite{singleAndDualArm}. Although well recognized for their scalability to high-dimensional problems, sampling-based approaches suffer from the narrow passage problem \cite{narrowPassageProblem} which dramatically inhibits their performance in our domain. The narrow passage problem refers to the poor performance of the sampling algorithms when the free space consists of narrow passages that have to be traversed to reach a goal. From the given model of the turbine we can be confident that there are several narrow passages that need to be navigated to reach regions of inspection. Optimization strategies seem like attractive options as they directly operate in the continuous space. However, using them to plan a long-horizon path for a hyper-redundant system like ours, in an environment riddled with constraints and minimas is extremely time-consuming. The highly cluttered environment in our domain provides excellent reasons for computing an informative heuristic function and using it to guide a search-based planning approach. 
% Hence, we adopted a search-based planning strategy. 
% which operates by discretizing the continuous action space into a small predefined set, which it then uses to construct a lattice graph that it searches over.

The performance of a search algorithm is significantly dependent on the discrete action space chosen and the heuristics used. While the choices for them are straightforward for more traditional problems, a domain such as ours requires careful reasoning. Due to the high clutter in the environment, the high dimensionality, and the joint limit constraints, direct applications of heuristics and action spaces from traditional manipulation domains drive the search into deep minima. Hence, our contributions in this paper are directed towards constructing and utilizing a more suitable action space and heuristics for our problem.

A search-based approach translates planning problems in continuous spaces to search problems in discrete graphs. 
% While searching the graph for a solution is a much faster affair, it comes at the expense of compromising the mobility of the system. 
The graph is constructed by discretizing the continuous action space into a finite set. 
% This restricts the states you can get to from any given state to a discrete set, which in reality is a continuous space. 
The effect of this discretization in most manipulation problems is insignificant. However, given the constrained space in our environment there are often situations where a simple predefined action set from a given state is unable to generate valid transitions that can help the search progress. 

% To address this, we formalize the task of neighbor generation as an optimization problem. This allows us to directly operate in the continuous space of the system overcoming the drawbacks of a discrete action space. Since the problem of generating a neighbor is significantly smaller when compared to solving the planning problem as a whole, the time required for this process is much lesser. At the same time, it is still far too high to be queried for every state encountered within a search (which is typically in the range of hundreds of thousands). 
% To combat this, we introduce optimization-based actions as actions that are invoked \emph{dynamically} on an as-needed basis in addition to the predefined action set (the set of static motion primitives) that is used throughout the search. Due to the computational expense associated with generating the optimization-based actions, they are invoked by the search only when it detects being stuck in a local minima. 
% % Hence, we use a predefined action set for the search. And when the inadequacy of the discrete set drives the search into a local minima, we solve an optimization problem to generate an action that can help the search out of the minima. 
% Inspired from the concept of lazy collision checking \cite{generalizedLazySearch}\cite{lazySP}, we further minimize queries to the optimizer by developing a lazy generation strategy. This allows us to postpone the actual call to the optimizer until absolutely needed.   

Our first contribution addresses this issue. We formalize the task of neighbor generation as an optimization problem. This allows us to directly operate in the continuous space of the system overcoming the drawbacks of a discrete action space. Since the problem of generating a neighbor is significantly smaller when compared to solving the planning problem as a whole, the time required for this process is much lesser. At the same time, it is still far too high to be queried for every state encountered within a search (which is typically in the range of hundreds of thousands). To combat this, we introduce optimization-based actions as actions that are invoked \emph{dynamically} on an as-needed basis in addition to the predefined action set (the set of static motion primitives) that is used throughout the search. Due to the computational expense associated with generating the optimization-based actions, they are invoked by the search only when it detects being stuck in a local minima. 

The concept of generating motion primitives online for manipulation problems was first introduced in \cite{singleAndDualArm}. They refer to such actions as Adaptive Motion Primitives and define them to be actions computed online and always leading to the goal configuration. These were designed to achieve precise positioning at goal configurations which do not necessarily lie on the discretized lattice. However, the goal of optimization-based action is different. It is to overcome the minima created by the discretization. Further, they do not necessarily lead to the goal and are queried at numerous stages within the search. 

Our second contribution is the development of a lazy generation strategy which is aimed at further minimizing queries to the optimizer. Inspired from the concept of lazy collision checking \cite{generalizedLazySearch}\cite{lazySP}, we have developed a lazy generation strategy which allows us to postpone the actual call to the optimizer for generating an action until absolutely needed. This way the search only queries the optimizer for generating actions that it intends to use (i.e., actions that are potentially part of the solution).

Our final contribution lies in the area of developing an informed heuristic for our problem. Using the solution of a lower-dimensional problem as a heuristic to guide the search in full dimensions is a common approach in the search literature. A good example of this would be to perform a 3D Breadth-First Search (BFS) in the workspace of a manipulator to guide a search in its configuration space \cite{singleAndDualArm}. While this is a good heuristic for common manipulation problems it forgoes information about the robot's reachability. A similar approach for our domain would be ineffective due to the extensive reachability constraints present in the system. 

Similar problems in other domains like tethered robot planning and humanoid footstep planning have been addressed by the use of topology-based heuristics \cite{homotopyTetheredRobot}\cite{homotopyHumanoid}. This class of heuristics reasons about the topological classes induced in the workspace by the obstacles present in the environment. Relevant topological classes for a given planning problem are identified and the search is guided along each of them simultaneously through the use of a Multi-Heuristic framework. We adopt a similar approach for our domain which we outline in the following sections.

\begin{figure}[t]
    \centering
    \includegraphics[width = 0.6\columnwidth]{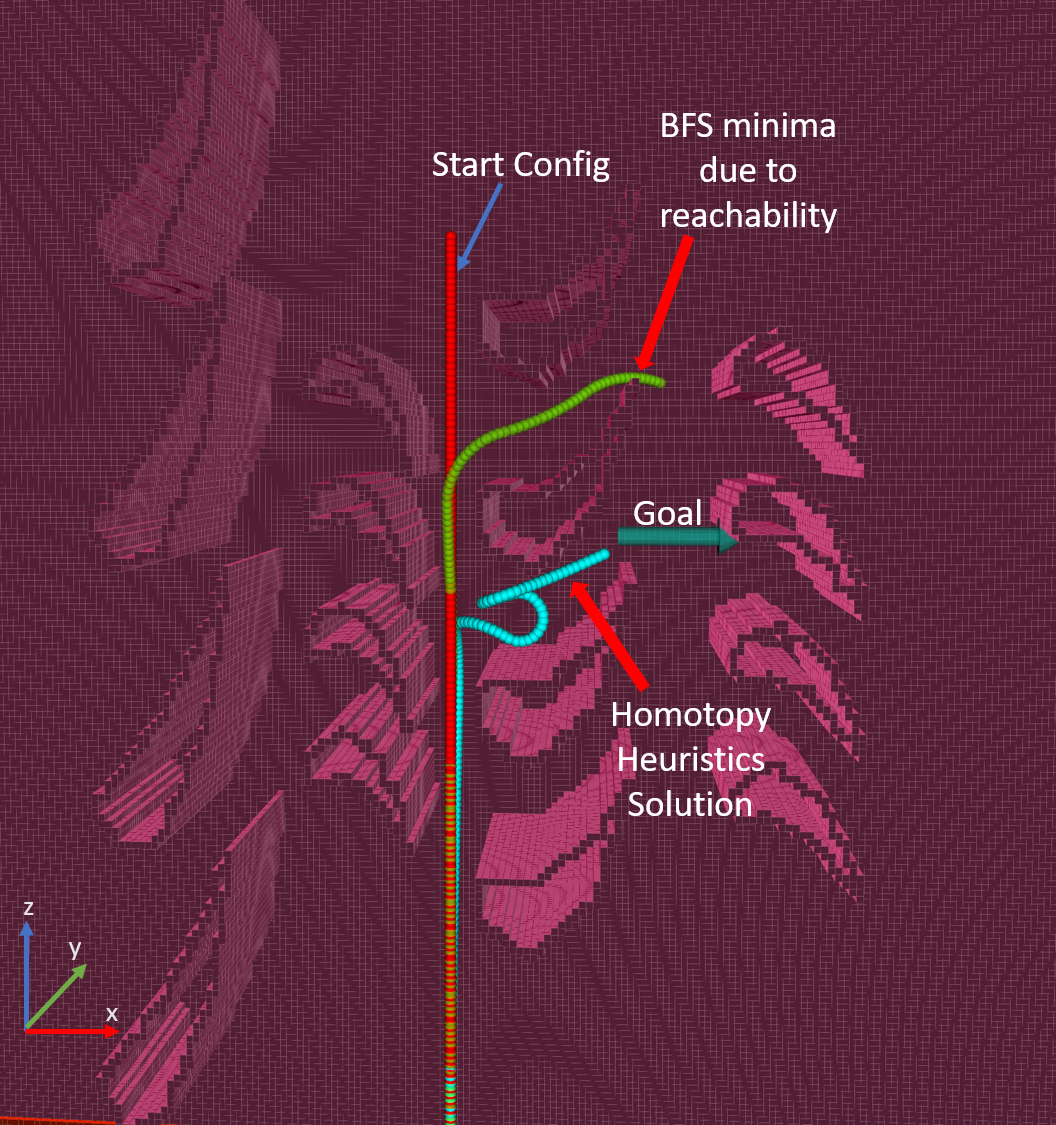}
    \caption{Minima caused by reachability constraints.}
    \label{fig:top_minima} \vspace{-0.2cm}
\end{figure}

\section{Preliminary}
\subsection{Search based Planning}
Our search based planning approach is rooted in A*. A* finds a path from a start state to a goal state, by repeatedly expanding nodes from a priority list referred to as OPEN. States in the OPEN list are prioritised based on the function $f(n) = g(n) + h(n)$. Here, $g$-value represents the cost of reaching the state $n$ from the start state and $h$-value otherwise referred to as the heuristic value, is an estimate of the cost of reaching the goal from $n$. The task of expanding a state entails computing the valid successors/neighbors that can be reached from the given state and adding them to the OPEN list based on their priority. Starting with the expansion of the start state, the search terminates when the goal state is expanded. Using the mentioned priority function for the OPEN list guarantees minimal computational effort (as measured by the number of state expansions) to find a provably optimal path to the goal provided the search uses a consistent heuristic. 
% A variant to A* which has shown great performance on high-dimensional problems is weighted A*. This uses a priority function of  $f(n) = g(n) + w * h(n)$ and is guaranteed to find a $w$-bounded suboptimal path.

\begin{figure}[!tbp]
  \centering
  \subfloat[]{\includegraphics[width=0.4\columnwidth]{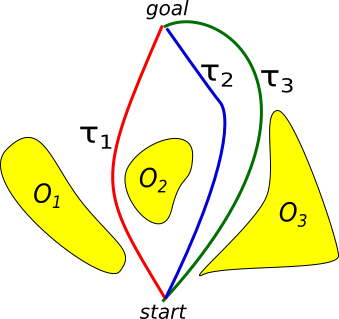}\label{fig:f1}}
  \hfill
  \subfloat[]{\includegraphics[width=0.5\columnwidth]{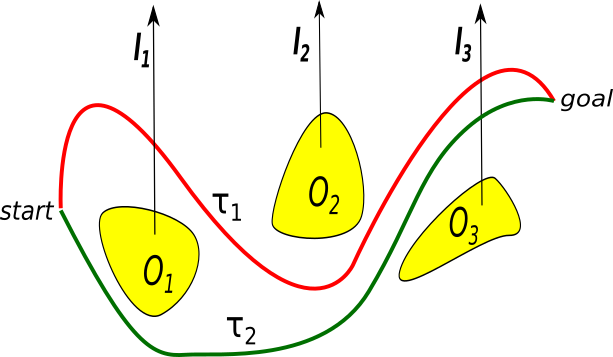}\label{fig:f2}}
  \caption{a) Curves $\tau_1$ and $\tau_2$ are in different homotopy classes. $\tau_2$ and $\tau_3$ are in the same homotopy class. b) h-signature of $\tau_1$ is $l_1 l_3$ and h-signature of $\tau_2$ is $l_3$.}
  \vspace{-0.5cm}
\end{figure}
\subsection{Homotopy Classes of Curves}
Two curves $\tau_1$ and $\tau_2$ connecting the same end points are said to be in the same homotopy class if they can be continuously transformed into one another without intersecting any obstacle present in the environment. Continuously deforming curves into one another to determine their respective homotopy classes is a highly non-trivial and expensive operation. This leads us to the functional H($\tau$) a homotopy invariant signature (h-signature) that can uniquely identify the homotopy class of a given curve. There have been different techniques developed over the years to compute this signature \cite{beamGraph} \cite{topologyBhattacharya}. As we reason about homotopy classes in 2-dimensions, we use the reduced word-based method developed in \cite{beamGraph}, owing to the simplicity of the approach for computing h-signatures of curves defined in 2-dimensional spaces. 

To compute the h-signature of a curve, we pick a point on each obstacle $\mathcal{O}_k \in \mathcal{O}$ in the environment and extend a beam $b_k$ vertically upwards from each of them (until the beam encounters either the boundary of the environment or the boundary of another obstacle $\mathcal{O}_i \in \mathcal{O}$, whichever comes first). We associate each beam $b_k$ with a unique letter $l_k$. To compute the h-signature of a curve $\tau$, we gather the letters of the beams crossed by the path as it progresses from start to goal (in order). If the curve intersects the beam $b_k$ from one direction (say left-to-right) the letter is recorded as $l_k$. And when the curve intersects the beam from the opposite direction (right-to-left) the letter is recorded as $\bar{l_k}$. To compute the final h-signature of the curve $\tau$ we reduce the word by removing continuous occurrences of any letter $l_k$ and its negation $\bar{l_k}$. This reduced word uniquely maps curves to their respective homotopy classes.
% Contributions: 
% \begin{itemize}
%     \item Optimization Actions
%     \item Topology Heuristics and a lazy way to incorporate them
% \end{itemize}
% This template provides authors with most of the formatting specifications needed for preparing electronic versions of their papers. All standard paper components have been specified for three reasons: (1) ease of use when formatting individual papers, (2) automatic compliance to electronic requirements that facilitate the concurrent or later production of electronic products, and (3) conformity of style throughout a conference proceedings. Margins, column widths, line spacing, and type styles are built-in; examples of the type styles are provided throughout this document and are identified in italic type, within parentheses, following the example. Some components, such as multi-leveled equations, graphics, and tables are not prescribed, although the various table text styles are provided. The formatter will need to create these components, incorporating the applicable criteria that follow.

\section{Problem Formulation}
\subsection{Robot Design}
The robot consists of ten flexible units attached to each other through a two-axis revolute joint, allowing the units to yaw and pitch about each other. The first unit is connected to a fixed base through a prismatic actuator which can propel the entire robot ahead. Therefore, the state-space $\mathcal{S}$ of the ten unit robot is given by $\{l \times \theta_{1} \times \phi_{1} \times \theta_{2} \times \phi_{2}..... \theta_{10} \times \phi_{10}\}$ where $l$ is the extension of the base prismatic joint and $\theta_{k}$ and $\phi_{k}$ are the pitch and yaw angles of $k^{th}$ unit with respect to the $k - 1^{th}$ unit. Hence, the total degrees of freedom of the system is 21. 

Each flexible unit is modeled using 15 rigid cylindrical subunits. The subunits are connected to each other through two axis revolute joints which mimic the joint controlling the unit. Hence, a pitch of $\theta_1$ in the state-space corresponds to the 15 subunits associated with unit 1 having a pitch of $\theta_1$ with respect to the previous subunit. Though there are 15 two axis revolute joints associated with each unit, since the joints mimic each other, the degrees of freedom associated with each unit remains 2. The actual hardware design and control of each flexible unit are beyond the scope of this paper. 

A camera with a highly constrained field of view is attached to the head of the robot i.e. the tip of the $10^{th}$ unit and is used for inspecting the turbine. 

\subsection{Problem Statement}
The overall goal of the domain is to use the snake robot-like manipulator to inspect regions of interest within a cluttered turbine. Given the environment, let $S_{valid}$ represent the space of all valid robot configurations - configurations that are not in collision with the turbine, not self-colliding, and within the joint limits. A transition $s_i \rightarrow s_{i+1}$ from state $s_i$ to state $s_{i+1}$ is said to be valid, if all the intermediate states $s_{int}$ lie in $\mathcal{S}_{valid}$. Since the controller of the system has been designed to linearly interpolate in the joint space, the intermediate states $s_{int}$ for a transition $s_i \rightarrow s_{i+1}$ can be obtained by linear interpolation. 

If $\pi$ represents a sequence of states $\{s_1, s_2 … s_t\}$  and $c(s_i, s_{i+1})$ represents the cost of transition $s_i \rightarrow s_{i+1}$, $C(\pi)$ represents the sum of the cost of all transitions in the path $\pi$.
Given a start state $s_{start} \in \mathcal{S}_{valid}$ and an end-effector goal pose $q_{goal} \in SE(3)$ within the turbine (from where you can view the regions of interest using the attached camera), the planning problem can be formally stated as to: 
\begin{equation}\label{eq:problem}
\begin{aligned}
    \text{find } &\pi^* = \argmin_\pi C(\pi) \\
    \text{s.t. } &s_i \in \mathcal{S}_{valid}, \,\forall\, s_i \in \pi \text{\qquad\qquad(path of valid states)} \nonumber \\
    &s_{int} \in \mathcal{S}_{valid}, \, \forall s_i \rightarrow s_{i+1} \in \pi \text{\quad(valid transitions)}\\ 
    &s_1 = s_{start}, s_t \in \mathcal{S}_G \text{\quad \qquad(start, goal constraints)} 
\end{aligned}
\end{equation}

Here $\mathcal{S}_G$ represents the goal set consisting of robot configurations for which the end-effector pose is at $q_{goal}$. For our problem, the cost of a transition $c(s_i, s_{i+1})$ is defined as the Euclidean distance between the end-effector positions ($\in \mathbb{R}^3$) of the states $s_{i}$ and $s_{i+1}$.  

\section{Optimization-based Action}
\begin{figure}[!tbp]
  \centering
  \subfloat[]{\includegraphics[width=0.45\columnwidth]{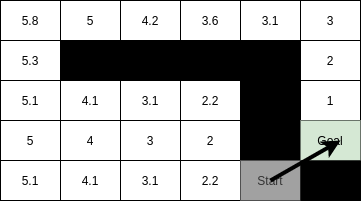}}
  \hfill
  \subfloat[]{\includegraphics[width=0.45\columnwidth]{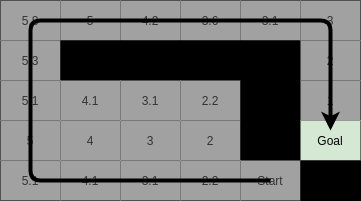}}
  \caption{a) Search terminates in a single expansion with the eight connected action space. b) Search has to expand all the states in the grid with the four connected action space}
  \label{fig:example_tabular}
  \vspace{-0.4cm}
\end{figure}
Let us consider the simple example shown in Fig. \ref{fig:example_tabular}. An agent has to get from the start cell to the goal cell as marked on the grid. Cells completely blacked out are occupied and cannot be transitioned into. The value in every cell represents the heuristic which is the Euclidean distance of the cell from the goal. If we run A* with an eight connected action set (the agent can transition to any of the immediate neighbor cells including diagonally opposite cells), this problem requires a single expansion. However, when the action set is restricted to four connected and no diagonal moves are permitted, it requires expanding every cell in the grid. 

The general discretized action space adopted for the manipulation domain is a set containing small increments and decrements of each joint \cite{singleAndDualArm}. For a 21-DOF manipulator, the size of this action set would be 42 (2*21). The idealogy behind this action set and the action set from the above example is the same. These are barebone primitive actions and combinations of them could result in more complex and useful motions. For example, in the above case the diagonal motion required to reach the goal could be achieved by moving one cell to the right and one cell upwards. However, both the action sets make the assumption that \emph{the barebone primitives result in valid transitions}. And this clearly went awry in the example scenario resulting in significantly more expansions. Since our environment consists of extremely narrow spaces and crevices that need to be navigated, the effect of this assumption is significant in comparison with common manipulation domains. 

\begin{figure}[!tbp]
    \centering
    \includegraphics[width = 0.65\columnwidth]{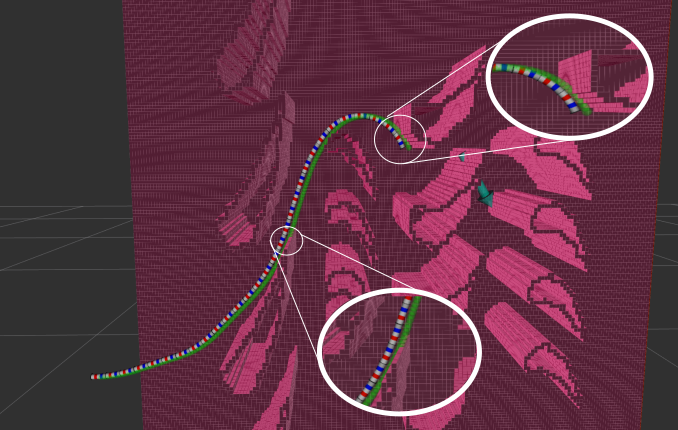}
    \caption{Example of a minima created by predefined action set. State in green represents a potential neighbor computed using the predefined action set.}
    \label{fig:opt_minima}
    \vspace{-0.3cm}
\end{figure}

An example of a minima created by the predefined action set can be seen in Fig. \ref{fig:opt_minima}. From the given configuration, there exists no action mentioned in the predefined set that helps the robot progress towards the goal. Similar to the state in green, any neighbor that progresses the end-effector towards the goal collides with the environment. Hence, the search spends a large amount of time expanding similar configurations.

A simple solution to this problem would be to add more complex actions to the predefined action set (combinations of the barebone actions). However, for our 21 dimensional system combining pairs of primitive movements alone would result in 420 actions ($2*21C_2$). This is despite disregarding weighting the primitives while combining them (for example an action could involve moving the first joint by 1 unit while moving the second joint by 3 units). Furthermore, given the structure of the robot and the environment, we would require combining more than pairs of primitives.
% actions combining more than pairs of primitives will be required. 

To counter this problem we propose an approach that combines the use of a (sparse) predefined set of actions with the use of optimization-based actions. The latter are constructed dynamically on an as-needed basis, specifically whenever the search encounters local minima, and utilize optimization to compute motions that propel the search out of the minima.
% we construct a predefined action set that we think would suit our problem. And when this action set drives the search into a minima, we solve an optimisation problem to generate a neighbor that would help the search out of the minima. 
Formalizing the generation of a neighbor as an optimization problem allows us to directly operate in the continuous space of all joint angles overcoming the issues encountered by the discrete action space. 

Minimas in the search are detected by observing heuristic stagnation \cite{heurStagnation}. That is, if the heuristic value of the states expanded does not improve over several expansions, it implies that the search is stuck in a local minima. Whenever this occurs, optimization-based actions are generated as additional actions available at the states being expanded.
% the optimizer is queried with the last expanded state.  

\subsection{Optimization Problem}
Let $s_{min}$ represent the state being expanded while the search is stuck in the local minimum. The goal of the optimization problem is to generate neighbors from $s_{min}$. Reasoning about the system in its configuration space is significantly more difficult than reasoning about it in the end-effector space of the robot. There is no direct way to identify configurations in the state space such that moving towards them would help us progress towards the goal. Hence, we choose to define the neighbors to $s_{min}$ in the end-effector space. This is a natural decision, since the goal of the overall problem is to cause meaningful motions to the end-effector which ultimately results in it reaching the defined goal pose.

% We consider primitive motions in the end-effector space. 
Since the end-effector space is $\mathbb{R}^3$, like any other 3D domain we can create a simple six connected action set (unit positive and negative motions along each axis). Although a unit motion in the end-effector space seems like a simple task, for the situations encountered in the search, this would require moving potentially all the joints of the robot simultaneously. Hence, for a given $s_{min}$ we solve six different optimization problems to compute potential neighbors. 

To define the optimization problem we need to identify an objective function that will enable reaching a 3D end-effector goal $g$ from a state $s$ (which in our case would be $s_{min}$). We formulated an objective that consists of three complementary components pertinent to our problem - i) Obstacle Cost, ii) Goal Cost, and iii) State Cost. 

\begin{equation}
\begin{aligned}
    \argmin_{s' \, \in \, \mathcal{S}} \mathcal{F}_{obj}(s, g, s') = \mathcal{F}_{obst}(s) + \lambda \mathcal{F}_{goal}(s, g) \\ + \, \gamma \mathcal{F}_{state}(s, s')
\end{aligned}
\end{equation}

The obstacle cost encourages staying away from the obstacles in the environment by penalizing states that are in collision. This component is very similar to the obstacle cost used previously in \cite{CHOMP}. It can be quantified as the inverse of the sum of the distances of every point on the robot from the nearest obstacle. Let $c_{obst}: \mathbb{R}^3 \rightarrow \mathbb{R}$, represent the obstacle workspace cost (which is equivalent to a distance field) and $fk : (s,u) \rightarrow \mathbb{R}^3$, be the forward kinematics function that returns the workspace position ($\in \mathbb{R}^3$) of a body element $u \in \mathcal{B}$ for a specific robot configuration $s \in \mathcal{S}$. Then the obstacle cost function can be expressed as:  
\begin{equation}
    \mathcal{F}_{obst}(s) = \frac{1}{\int_{\mathcal{B}} c_{obst}(fk(s,u)) du}
\end{equation}

The role of the goal cost is to encourage moving towards the goal $g$. It is defined as the Euclidean distance in 3D between $g$ and the end-effector pose of the potential neighbor $s’$. If $u_{ee}$ represents the end-effector
\begin{equation}
    \mathcal{F}_{goal}(s', g) = Euclidean(fk(s', u_{ee}), g) 
\end{equation}

Since the goal for the optimization problem is defined in the end effector space it is important to ensure that the robot does not have to reconfigure itself severely to reach the goal. For this purpose we define the third term of our objective, the state cost. This measures the Euclidean distance between the state in the minima $s_{min}$ and the potential neighbor $s'$ in the configuration space. The obstacle cost does not explicitly reason about the validity of the transition itself. Rather, it only tries to ensure that the generated neighbor is collision free. However, this additional term ensures that the reconfiguration is minimal thereby increasing the chances of a valid transition. 
\begin{equation}
    \mathcal{F}_{state}(s, s') = Euclidean(s, s')
\end{equation}

The generated neighbors are checked for collisions and valid transitions from $s$ before being added as valid successors in the search.

\begin{algorithm}[t]
\begin{small}
\caption{\textsc{Lazy Generation based Search}}\label{alg:LGA*}
% % \textbf{Input:} Planning problem $\PP$, number of \AMP subgoals $N$, number of \AMP samples $M$, simulation start time $t_\text{sim}$, planning timeout $t_\text{max}$ \\
% % \textbf{Output:} solution path $\pi$
\begin{algorithmic}[1]
\Procedure{Plan}{$s_{start}, \mathcal{S}_G$}
    \State $start.state = s_{start}$
    \State $start.g = 0;$  \hspace{0.2cm} $start.h = \texttt{Heuristic}(start.state)$ 
    \State $\texttt{Insert}$($OPEN$, $start$) 
    \While{$OPEN$ is not empty}
        \State remove $s$ with smallest $f(s)$ from $OPEN$
        \If{\texttt{pseudostate}($s.state$)}
            \State $\hat{s}$ = $\texttt{queryOptimizer}(s.parent.state$, $s.state$) \label{alg:line:generatestate}
            \State \Comment{Generating state corresp. to end effector goal}
            % \State \Comment{Returns state corresponding pseudostate}
            \If{$\texttt{validTransition}(s.parent.state, \hat{s})$} \label{alg:line:start_reevaluation}
                \State $s.state$ = $\hat{s}$ 
                \State $s.g = s.parent.g + c(s.parent.state, s.state)$
                \State $\texttt{Insert}(OPEN, s)$ \Comment{Reinsert into $OPEN$}
            \EndIf
        \State continue \label{alg:line:end_reevaluation}
        \EndIf
        \If{$s.state$ in $CLOSED$}
            \State continue
        \ElsIf{$s.state \in \mathcal{S}_G$}
            \Return \texttt{ExtractPath}($s$)
        \Else
            \State \textsc{Expand}(s)
        \EndIf
    \EndWhile
\EndProcedure
\Procedure{Expand}{$s$}
    \State $\texttt{Insert}$($CLOSED$, $s.state$)
    \State $S' = \texttt{getSuccessors}(s.state)$ \Comment{Predefined action set}
    \If{\texttt{localMinima}} \label{alg:line:minima}
        \State $S' += \texttt{getOptPseudoSuccessors}(s.state)$ \label{alg:line:start_pseudoinsertion}
        \State \Comment{Returns 3D end effector goals}
    \EndIf
    \For{$s' \in S'$}
        \If {not $\texttt{pseudostate}(s')$}
            \If {$\texttt{validTransition}(s.state, s')$} 
                \State $succ.g = s.g + c(s.state, s')$
            \Else
                \State continue
            \EndIf
        \Else
            \State $succ.g = s.g + \bar{c}(s.state, s')$ \label{alg:line:optimistic_cost}
            \State \Comment{Underestimate of true cost}
        \EndIf
        \State $succ.state = s'$; \hspace{0.05cm} $succ.parent = s$ 
        \State $succ.h = \texttt{Heuristic}(succ.state)$
        \State $\texttt{Insert}(OPEN, succ)$ \label{alg:line:end_pseudoinsertion}
    \EndFor
\EndProcedure 
\end{algorithmic}
\end{small} 
\end{algorithm} 

\subsection{Lazy Generation}
Lazy search algorithms are a class of search algorithms developed for domains where the task of edge (transition) evaluation is computationally expensive, thereby acting as the major bottleneck for the planning problem \cite{generalizedLazySearch}\cite{lazySP}. These approaches vie to only evaluate edges that are essential for the search. A good example from this class of algorithms would be Lazy Weighted A* (LWA*) \cite{LWA*}. This algorithm postpones evaluating edges until expansion. This way the algorithm only evaluates edges to states that are potentially part of the (bounded sub-)optimal solution. 
% Meaning an edge to a state is evaluated only if the state is expanded by the search. If the edge is valid, the state is expanded and the search proceeds. If the edge is invalid, the state is discarded and the edge connecting to the next state in the OPEN list is evaluated. 

However, the optimization primitives provide us with the unique opportunity of exploring a different dimension of lazy algorithms. The bottleneck in this problem is the generation of the optimization primitive rather than collision checking an existing transition (each generation approximately takes 1.5 seconds). Drawing inspiration from LWA* we choose to delay the generation of the primitive until expansion. Thereby, we query the optimizer only for states essential to the problem. 

An outline of the lazy generation based search can be found in Alg. \ref{alg:LGA*}.
When a local minima in the search is detected, optimization-based actions are queried as additional actions from the state $s$ being expanded (Line \ref{alg:line:minima}). Instead of solving the optimization problems right away and adding the solution states as successors into the OPEN list, we create pseudostates representing the solutions of the optimizer that we add to the OPEN list (Lines \ref{alg:line:start_pseudoinsertion}-\ref{alg:line:end_pseudoinsertion}). An optimistic estimate of the cost of transition $\bar{c}$, from $s$ to the potential successor $s'$ is used to compute the $g$-value of the pseudostate (Line \ref{alg:line:optimistic_cost}). Optimistic cost estimate implies that $\bar{c}(s,s’) \leq c(s,s’)$. Only when the pseudostate is selected for expansion, we generate the full configuration of this successor (Line \ref{alg:line:generatestate}). If the transition to this configuration from $s$ is invalid, we discard the state and proceed with expanding other states in the OPEN list. If the transition is valid, then we now know the edge’s true cost, so we reinsert it into the OPEN list (Lines \ref{alg:line:start_reevaluation}-\ref{alg:line:end_reevaluation}). And when it is popped from the OPEN list the second time, it will actually be expanded and will generate successors of its own. Lazy generation based search similar to LWA* is both complete and bounded suboptimal. 

In our problem domain, both the cost function and the heuristic function are computed in the end-effector space. Hence, both the $g$-value and the $h$-value of the state can be computed from the end-effector position of the potential neighbor $s’$ (obtained through the 6 connected grid described in the previous subsection). Hence, there is no requirement for an optimistic estimate of the transition cost. 

\section{Topology Heuristics}
% The obstacles present in the environment introduce geometric and topological constraints for the robot.  Meaning that some goals in the robot workspace can only be reached if the path lies in specific topological classes. The joint limits present in the system (specifically on the prismatic joint), create constraints on the robot’s reachability making the goals unreachable through the other classes. Not reasoning about these constraints during the search could lead to scenarios similar to Fig. \ref{fig:top_minima}, where the search is directed toward a minima created by the constraint. This calls for a more informed heuristic that reasons about the topology classes induced by the blades of the turbine. 

% In our domain, the limited room along the $y$-axis of the turbine forces the robot to primarily navigate around the obstacles in the $x-z$ plane (refer Fig. \ref{fig:top_minima} for axes). Therefore, it is sufficient to guide the search along the topological classes created by the projection of the turbine onto the $x-z$ plane. Choosing a desired homotopy class for a planning problem corresponds to choosing the homotopy class of the 2-D projection of the curve corresponding to the goal configuration. The curves in Fig. \ref{fig:f2} in the context of our problem would correspond to the projection of the final configuration of the snake robot-like manipulator. 

The invariance of the turbine environment along the $y$-axis (refer Fig. \ref{fig:top_minima} for axes) and the majority of the robot’s transitions lying in the $x-z$ plane (sideview plane of the turbine) make it sufficient to discuss about topology classes in the 2-D $x-z$ plane. Choosing a desired homotopy class for a planning problem corresponds to choosing the homotopy class of the 2-D projection of the curve corresponding to the goal configuration. The curves in Fig. \ref{fig:f2} in the context of our problem, correspond to the projection of the final configuration of the snake robot-like manipulator. 

The obstacles present in the environment and the joint limits present in the system (specifically the prismatic joint) introduce reachability and topological constraints for the robot. Meaning that some goals in the robot workspace can only be reached through specific homotopy classes. Not reasoning about these constraints during the search could lead to scenarios similar to Fig. \ref{fig:top_minima}, where the search is directed toward a minima created by the constraint. This calls for a more informed heuristic that reasons about the homotopy classes induced by the blades of the turbine. 

The role of the homotopy heuristic is to estimate the cost of reaching the goal pose from any state $s$ through a specific homotopy class in the projected space. In common manipulation domains when the goal is to compute a heuristic from a state, the distance of the end-effector position of the state $s$ from the goal pose is estimated. The task here is to estimate the cost of reaching the goal through a specific homotopy class. Hence, the distance of the end-effector of the state to the goal pose through the class of interest is estimated.

Consider the example shown in Fig. 6. Let the curve in red represent the 2-D projection of a state (full robot configuration) for which the heuristic has to be computed. As can be seen in the figure, the curve corresponds to an h-signature of $l_1$. Hence, if the desired signature for the final configuration is $l_1$, we can estimate the end-effector distance to the goal along the " " (empty) signature and use this as the heuristic (which in this case would be $d_1$). If the desired signature for the final configuration is $l_1l_2$ we can estimate the end-effector distance along the class $l_2$ (which in this case would be $d_2$). 

For any state $s$ encountered in the search, the homotopy heuristic is obtained by computing the distance between the 2D projection of the end effector position of the state $s_{proj}$ and the projection of the goal pose $g_{proj}$, through a homotopy class $h$. $h$ can be obtained by subtracting the signature of the state $s$ from the desired goal configuration signature. In the example mentioned above, we subtracted $l_1$ from the desired signature of $l_1l_2$ and computed the distance along $l_2$ class. These distances can be obtained by querying the homotopy augmented graph discussed below.
\vspace{-.2cm}
\begin{figure}[t]
    \centering
    \includegraphics[width = 0.8\columnwidth]{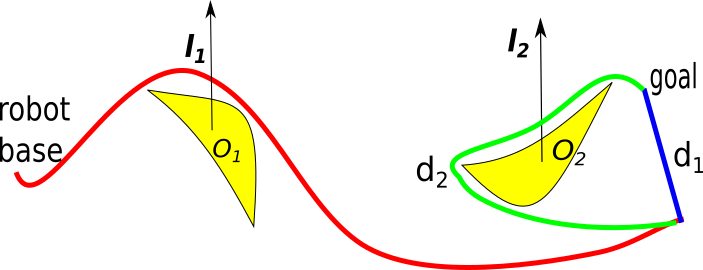}
    \caption{The curve in red represents the 2D projection of a robot state. $d_1$ estimates the distance of reaching the goal when the desired class is $l_1$ and $d_2$ estimates the distance of reaching the goal when the desired class is $l_1l_2$.}
    \label{fig:passage}
    % \vspace{-0.5cm}?
\end{figure}

\subsection{Homotopy Augmented Graph}
% The limited room along the $y$-axis of the turbine forces the robot to primarily navigate around the obstacles in the $x-z$ plane (refer Fig. \ref{fig:top_minima} for axes). Therefore, it is sufficient to guide the search along the topological classes created by the projection of the turbine onto the $x-z$ plane. 

% When the goal is to compute a heuristic that can estimate the distance from any state to the goal, we generally perform a 3-Dimensional Breadth First Search (BFS) from the 3D goal position to every other point in the robot's workspace. This distance in the robot's workspace is used as a heuristic to guide the full dimensional search. 
The goal of this routine is to estimate the distance from points in the 2D projection of the robot workspace to $g_{proj}$ through specific homotopy classes. For this purpose we do a BFS in a homotopy augmented graph $G_h$  \cite{beamGraph}\cite{topologyBhattacharya} from the node ($g_{proj}$," "), where " " corresponds to the empty signature. Each vertex in this graph consists of the 2D projection of a point in the workspace and the corresponding h-signature of the path leading up to the point. This additional state variable keeps track of the homotopy of the path from the goal to the current point. Therefore, the distance from any node $(v, h)$ to the goal node (computed by the BFS), represents the distance from $v$ to $g_{proj}$ measured through the homotopy class $h$. 
% Since the size of this graph could quickly explode, we prune away nodes with signatures that are not subsets of the desired signature. 
% Meaning, if the desired signature is $l_1l_2$ only nodes with signature " ", "$l_1$", "$l_1l_2$" are computed. 

% It is also to be mentioned that even if we would like to extract the distance along different topological classes, a single search in this graph $G_h$ is sufficient as the graph has multiple nodes for each point $v$ in the projected workspace, $\{v,h_1\}$, $\{v,h_2\}$... where each node represents reaching the point through different homotopy classes. 

% As the size of this graph $G_h$ can quickly explode, we perform a few optimizations. Firstly, we prune away states whose h-signature suggests that the homotopy class of the paths might not be relevant to us. If the h-signature of interest is \textit{"ab"}, we only keep vertices whose h-signatures are substrings of the word \textit{(" ", "a", "ab")}. 
% Secondly, we prune away states whose distance from the goal is higher than the maximum reachable distance of the robot (fixed base constraint). Finally, we prune away states whose h-signatures correspond to looping around an obstacle. Paths that loop around obstacles are generally suboptimal and not preferred for our domain. 

% To compute the heuristic for a state $S$ being expanded in this queue, we compute the h-signature of the state $S$ by tracking the beam crossings of the 2-D projection of the path traversed by the end effector. 

\subsection{Detection of Relevant Homotopy Classes}
\begin{figure}[t]
    \centering
    \includegraphics[width = 0.65\columnwidth]{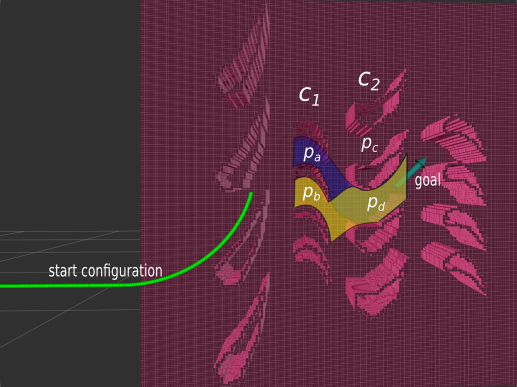}
    \caption{The goal can be reached from the given start configuration through different passage sequences $p_a p_d$ and $p_b p_d$.}
    \label{fig:passage}
    \vspace{-0.5cm}
\end{figure}
% One can easily identify relevant topology classes for a problem and manually input it into the algorithm. However, we employ an auto-detection scheme. This routine is highly specific to the turbine environment. The design of the turbine restricts the point of entry of the robot to a specific configuration as indicated in Fig. 1. And the topology detection scheme was devised while accounting for this information. As can be seen in Fig. 1 and 2, the blades of the turbine have been arranged into rows and columns. Consecutive blades in each column define passages through which the robot can traverse. Given the robot start configuration, the robot has to navigate the columns of the turbine from left to right. We number the columns from 0 to $n$ along the same direction.  

% We identify a point of entry and point of exit for each of the passages. Given this information, it is easy to identify passages in a column $c$ that are closest to the entry point of a specific passage in the next column $c+1$ and can be used to reach it. Hence, given a goal, we identify passages ${p_1, p_2..}$ closest to this goal and find sequences of passages in previous columns that can lead to the identified set ${p_1, p_2..}$. This way we can find sequences of passages (homotopy classes) relevant to our problem. 
The obstacles present in the turbine induce multiple topology classes for paths in the robot workspace. We observe that not all of these classes are relevant for a given planning problem. We developed a routine highly specific to the turbine environment that identifies relevant classes. It is also possible to pick these classes through manual inspection.
% Our approach identifies only the relevant obstacles in a scene and guides the search along all the classes induced by them. 

% Given the turbine environment, there are several topology classes induced by the blades and it is infeasible to guide the search along each of them. Hence, we developed a routine that is highly specific to the turbine to identify classes that are relevant to the problem. 

As can be seen in Fig. \ref{fig:passage}, the blades of the turbine have been arranged into rows and columns. Consecutive blades in each column define passages through which the robot can traverse. Since it is the obstacles that both induce the homotopy classes and create passages, the problem of identifying a homotopy class of interest can be translated into the problem of identifying a sequence of passages that can be traversed by the robot to reach the goal. 

Given a passage $p$ in a column, it is easy to identify passages in neighboring columns that are closest to $p$ and can be traversed through, to reach $p$. For example, as can be seen in Fig. \ref{fig:passage}, $p_d$ in column 2 is reachable through $p_a$ and $p_b$ in column 1. Given the start configuration and goal pose, we identify the columns of blades $\{c_1, c_2 .. c_n\}$ that need to be traversed by the robot to reach the goal. In the column nearest to the goal pose $c_n$, we identify the passage $p$ closest to the goal that can be used to reach the pose ($p_d$ in Fig. \ref{fig:passage}). Once we identify such a passage, we can compute the passage in the previous column $c_{n-1}$ that can be used to reach $p$ ($p_b$ or $p_a$). We repeat this process to identify the passages in all the relevant columns. This sequence of identified passages defines the homotopy class of interest. 

Since a passage in any column $c_n$ can be reached through multiple passages in $c_{n-1}$ it is possible to identify multiple homotopy classes that could be pursued. It is possible to prioritize sequences based on how close the passages are to each other. For example, $p_b$ is closer to $p_d$ in comparison with $p_a$. This means that the robot has to travel lesser distance to reach the goal. Thereby, it reduces chances of the search getting stuck in reachability created minimas. Thus, we prefer $p_bp_d$ over $p_ap_d$. 

\subsection{Search}

Since there could be multiple promising homotopy classes that the search could be guided along, we use MHA*, a Multi-Heuristic search framework \cite{mha*}. This way even if there exists a minima along one of the desired homotopy classes, we could find a solution along the others. MHA* makes use of multiple OPEN lists, with each list dedicated to a different heuristic function. In our case, this would mean that MHA* maintains individual OPEN lists for every homotopy class of interest. MHA* generally employs a round-robin scheduling strategy to choose the OPEN list to expand from, meaning that the search along every homotopy class would be given equal preference. We chose to employ a Dynamic Thompson Sampler (DTS) similar to \cite{DTS} which would give us the ability to better utilize resources. The DTS approach expands more often from OPEN lists that are more promising and making advances toward the goal. If the search along a specific homotopy class gets stuck in a minima, DTS allocates lesser resources to it and more resources towards a search that is making progress. 

\section{Results}
\begin{table}
\centering
\begin{tabular}{ P{2.1cm} P{1.6cm} P{1.6cm} P{1.6cm}  }
\toprule
\textbf{Metric} & \textbf{RRT-Opt} & \textbf{RRT-Search} & \textbf{Ours}\\
\hline 
Success Rate (\%) & 20.0 & 36.6 & \textbf{83.3} \\ \hline
Planning Time (s) & 277.2 & 292.3 & \textbf{167.0}\\
\bottomrule
\end{tabular}
\caption{Average performance of our approach in comparison with sampling based baselines.}\label{table:baselines}
\end{table}

\begin{table}
\centering
\begin{tabular}{ P{2.1cm} P{1.6cm} P{1.6cm} P{1.6cm}}
\toprule
\textbf{Metric} & \textbf{Predefined} & \textbf{Predefined} & \textbf{Predefined}\\
 & \textbf{+} & \textbf{+} & \textbf{Only}\\
 & \textbf{Lazy Opt} & \textbf{Opt} &\\
% \textbf{Metric} & \textbf{Predefined \newline \hspace{2cm} + \newline Lazy Opt} & \textbf{Predefined \newline + \newline Opt} & \textbf{Predefined Only}\\
\hline 
Success Rate (\%) & \textbf{83.3} & 60.0 & 20.0\\ \hline
Planning Time (s) & \textbf{167.0} & 233.7 & 300.4\\
\bottomrule
\end{tabular}
\caption{Impact of optimization action.} \label{table:action}
\end{table}

\begin{table}
\begin{tabular}{ P{2.1cm} P{1.2cm} P{1.2cm} P{1.2cm} P{1.0cm}}
\toprule
\textbf{Metric} & \textbf{Two} & \textbf{Two} & \textbf{One} & \textbf{BFS}\\
 & \textbf{Homotopy} & \textbf{Homotopy} & \textbf{Homotopy} & \\
 & \textbf{with DTS} &  \textbf{w/o DTS} &  & \\
\hline 
Success Rate (\%) & \textbf{83.3} & 73.3 & 66.6 & 23.3\\ \hline
Planning Time (s) & 173.9 & 258.1 & 197.6 & \textbf{134.7}\\
\bottomrule
\end{tabular}
\caption{Impact of homotopy heuristics and DTS.} \label{table:heuristic} 
\end{table}

In this section we present the performance of our algorithm through comparisons with sampling based baselines and through ablation studies which help isolate the impact of each component of our algorithm. A timeout of 1000 seconds was given for each of the studies, and a failure was recorded if an algorithm was unable to find a solution within the timeout. The results have been averaged over 30 different planning problems. The algorithm used for solving the optimization problem formulated in section IV A as part of our approach, is the stochastic and derivative free black box method of Covariance matrix adaptation evolution strategy (CMA-ES) \cite{CMA-ES}. Unless explicitly stated, variants of our approach were guided along the top two homotopy classes identified for the problem using techniques described in the previous section.

Table \ref{table:baselines} is a comparative study of our approach against two different versions of RRT \cite{RRT}. Due to the complexity of the problem, existing baselines like RRT do not work well when directly applied to the problem. Hence, appropriate modifications were made. The sampling for both the RRT baselines in Table \ref{table:baselines} is performed in the 3D workspace of the robot. The difference between the two variants is in the EXTEND routine of RRT which is responsible for extending the nearest neighbor in the tree towards a random sample. \subsubsection{\textbf{RRT-Opt}} Makes use of the optimizer to implement the EXTEND routine (the same optimizer as used in our own approach). Given a randomly sampled 3D point, it identifies an intermediate point close to the end-effector position of the neighbor and in the direction of the sampled point. It then proceeds to solve the optimization problem we had formulated in section IV A with the intermediate point as the goal and nearest neighbor from the tree as the state from which the neighbor has to be computed ($s$ in Eqn. 1). If the transition from the neighbor state and the solution of the optimizer has been validated, the process is repeated with a goal closer to the sampled point and the solution of the previous iteration as $s$. \subsubsection{\textbf{RRT-Search}} This variant employs a Weighted A* search algorithm to extend between the sampled point and the neighbor configuration. While utilizing the randomness associated with sampling based approaches, the baseline taps into the effectiveness of search-based techniques as well. We used a simple Euclidean heuristic and an action space similar to the one we had developed for our planner in section IV (including the optimization queries). A timeout of 7 seconds was given for every search query and the state with the minimum heuristic is returned. 

Computing inverse kinematic solutions is an expensive process for our 21-DOF system. And the cluttered environment makes it infeasible to sample collision free inverse kinematics solutions from the goal pose. This ruled out the possibility of bidirectional baselines like RRT-Connect \cite{RRT-Connect}. 

From Table \ref{table:baselines} we see that our approach significantly outperforms the sampling-based baselines. The baselines could solve not more than 36\% of the test cases in comparison with the 83\% success rate of our approach. The poor performances of the baselines can be primarily attributed to their narrow passage problem discussed in Section I. Outperforming our nearest competitor by solving 50\% more problems highlights the significance of our contributions.
% This table highlights the importance of our contribution as we have outperformed the nearest competitor by solving over 50\% more problems.

Table II and Table III present ablation studies associated with the optimization-based action and topology heuristics respectively. As can be seen in Table II, using just a predefined action set yields a poor success rate. However, the incorporation of the optimization action improves the performance of the planner several fold. It is also visible that the lazy generation technique is a crucial component of our algorithm improving success rate by over 20\%.

Table III clearly shows the benefit of using homotopy heuristics over a BFS heuristic. A significant improvement in success rate is observed. It also brings out the trade-off associated with using multiple homotopy heuristics. While having more heuristics slows the search down (as the search has to be guided in multiple directions simultaneously), it increases the overall success rate of the algorithm (as the more relevant directions you guide the search in, the more likely you are to reach the goal without getting stuck in a minima). The table also elicits the impact of using DTS over a round-robin scheduling strategy in the Multi-Heuristic framework.

% The table also brings out the trade-off associated with using multiple homotopy heuristics. While having more heuristics slows the search down, it increases the overall success rate of the algorithm allowing you to solve more problems. The benefit of DTS can also be observed from the table. Using DTS improves planning times by over 30\%.  

% Table II and Table III elicit the impact of topology heuristics and the optimization-based action space. We can clearly observe that the optimizer action is crucial for the overall performance of our algorithm. And the use of topology heuristics has increased success rate by over 80\% and a three-fold speed up in planning times. Table IV brings out the performance of lazy generation. 

\section{Conclusion}
In this paper the planning problem for a 21 DOF snake-robot like manipulator navigating a cluttered gas turbine was successfully tackled. The minima created by the inadequacy of a discrete action space in the highly constrained turbine environment was tackled through the development of optimization-based actions. Due to the computational expense associated with solving the optimization problem, a lazy generation technique was developed that delays the actual call to the optimizer until absolutely required. The topological constraints created by the blades of the turbine environment and the joint limits, was tackled through the development of homotopy heuristics that reason about the homotopy classes induced in the robot's workspace.  

\bibliography{refs}
\bibliographystyle{IEEEtran}

\end{document}